\documentclass[letterpaper, 10 pt, conference]{ieeeconf}  

\IEEEoverridecommandlockouts                              

\usepackage{url} 
\usepackage{graphicx} 
\usepackage{multicol} 
\usepackage{footmisc} 
\usepackage{subfigure} 
\usepackage{ctable} 
\usepackage{makecell} 
\usepackage{etoolbox}
\usepackage[T1]{fontenc} 
\usepackage{amsfonts} 
\usepackage{verbatim} 
\usepackage[all]{nowidow} 

\usepackage[ruled,vlined]{algorithm2e} 

\title{\LARGE \bf
What Happens When Pneu-Net Soft Robotic Actuators Get Fatigued?
}

\author{Jacqueline Libby, \textit{Member, IEEE}$^{1}$, Aniket A. Somwanshi$^{2}$, Federico Stancati$^{2,3}$, Gayatri Tyagi$^{4}$, Aadit Patel$^{2}$,
\\ Naigam Bhatt$^{2}$, JohnRoss Rizzo$^{1,5}$, S. Farokh Atashzar, \textit{Senior Member, IEEE}$^{1,2,6,7}$
\thanks{\hspace{-2.2em}$^{1}$Center for Urban Science and Progress, New York University (NYU), Brooklyn, NY 11201 USA.
        {\tt\small jlibby@nyu.edu}}%
\thanks{\hspace{-2.2em}$^{2}$Department of Mechanical and Aerospace Engineering, NYU, Brooklyn, NY 11201 USA.}
\thanks{\hspace{-2.2em}$^{3}$Department of Mechanical and Aerospace Engineering, Sapienza University of Rome, 00185 Rome, Italy.} 
\thanks{\hspace{-2.2em}$^{4}$Department of Computer Science and Engineering, NYU, Brooklyn, NY 11201 USA.}
\thanks{\hspace{-2.2em}$^{5}$School of Medicine, NYU, New York, NY, 10016}
\thanks{\hspace{-2.2em}$^{6}$Department of Electrical and Computer Engineering, NYU, Brooklyn, NY 11201 USA.}
\thanks{\hspace{-2.2em}$^{7}$\tt\small Corresponding author: f.atashzar@nyu.edu}%
}

\begin{document}

\maketitle
\thispagestyle{empty}
\pagestyle{empty}

\begin{abstract}
Soft actuators have attracted a great deal of interest in the context of rehabilitative and assistive robots for increasing safety and lowering costs as compared to rigid-body robotic systems. During actuation, soft actuators experience high levels of deformation, which can lead to microscale fractures in their elastomeric structure, which fatigues the system over time and eventually leads to macroscale damages and eventually failure.
This paper reports finite element modeling (FEM) of pneu-nets at high angles, along with repetitive experimentation at high deformation rates, in order to study the effect and behavior of fatigue in soft robotic actuators, which would result in deviation from the ideal behavior.
Comparing the FEM model and experimental data, we show that FEM can model the performance of the actuator \emph{before} fatigue to a bending angle of 167$^{\circ}$ with $\sim$96\% accuracy.
We also show that the FEM model performance will drop to 80\% due to fatigue after repetitive high-angle bending.
The results of this paper objectively highlight the emergence of fatigue over cyclic activation of the system and the resulting deviation from the computational FEM model. Such behavior can be considered in future controllers to adapt the system with time-variable and non-autonomous response dynamics of soft robots.
\end{abstract}

\section{Introduction}\label{sec.introduction}

With an aging worldwide population~\cite{kanasi2016aging}, the demand for accessible rehabilitation and physical therapy is increasing rapidly.
The intensity, duration, frequency, and timing of rehabilitative exercises are critical to recovery~\cite{han2008stroke},~\cite{takahashi2008robot},~\cite{page2012optimizing}.
For stroke patients, this therapy should be delivered in a timely manner since rehabilitation is most effective in the acute	stages~\cite{brocklehurst1978much},~\cite{paolucci2000early}. 
These factors put a high strain on our healthcare system, which calls for the development of in-home therapeutic systems.
During the COVID-19 pandemic, the need became pronounced due to concerns of infection spread. 
In-home rehabilitative robotic devices that can physically interact with patients in an intelligent and safe manner have the potential to address the aforementioned challenges~\cite{atashzar2021can},~\cite{centers2003public},~\cite{gresham2004post}. 

Rehabilitative technologies such as smart exercise machines and robotic exoskeletons are intended to help improve patient mobility, motor control, and strength by either assisting a patient through different rehabilitative motions or providing appropriate resistance during an exercise~\cite{maciejasz2014survey},~\cite{agrawal2018encyclopedia}.
In-home	robotic rehabilitation devices can help address the labor demand and allow for improved recovery~\cite{Stein2012-AmJPhysMedRehab}.
For patients to use rehabilitative technology in their homes, it should be safe, comfortable, and economically accessible, which soft robotics affords~\cite{polygerinos2015soft}. 
The inherent compliance of soft robots makes them safer for independent in-home use by patients in assistive and rehabilitative applications~\cite{cianchetti2018biomedical},~\cite{heo2012current}.
Made out of relatively inexpensive materials, soft robots are also more affordable when compared with rigid-body systems.

One of the fundamental research challenges in soft robotics is that soft materials change in behavior over time due to fatigue~\cite{miron2016design},~\cite{villegas2012third}.
In this work, we characterize fatigue for one of the most common types of soft actuators: the pneu-net.
Pneu-nets are pneumatically-operated soft actuators made from elastomeric materials with a network of air chambers.
They actuate in a bending motion due to the embedding of a strain-limiting layer.
The actuation will result in a rotation-like deformation, in a direction toward the strain-limiting layer and away from the regions of high strain.
The regions of highest stress/strain will undergo the most fatigue~\cite{terryn2021review}, which can result in micro-fractures (changing the behavior) or macro-fractures (breaking of the walls and leading to failure of the robot).
In the case of a pneu-net, these highest levels of fatigue occur in the upper regions of the interior walls between air chambers (see Fig.~\ref{fig.fem}).
Some have tried to optimize the design by discarding material from these high-stress/strain regions.
In~\cite{Mosadegh2014-AdvFunctMater}, the fast pneu-net (fPN) design was introduced as an improvement to reduce fatigue.
The fPN design involves cutting out the interior walls from the outside, creating a ridged external surface.
As another alteration, described in~\cite{Zhou2019-SaeIntJMaterManuf}, the interior walls are cut out from the inside by creating trapezoidal-shaped air chambers, allowing the design to maintain an almost continuous external surface.
Whether or not the design is optimized or constrained for a specific application, the elastomeric material of a soft actuator will still undergo strain, which will eventually lead to fatigue.
In this paper, we analyze this fatigue behavior, which is a necessary consideration for any soft robotics application, and show how it causes deviation from the FEM model.

It should be highlighted that the high strain within the interior walls of the sPN not only leads to fatigue but it also makes finite element modeling (FEM) a challenge.
Most FEM techniques have been developed for analyzing solid structures, such as metals with much less deformation than hyperelastic silicone, a common material used for soft actuators.
Thus, studies of soft actuators with high strain rates such as sPNs do not quantitatively compare FEM and experimental data. These studies either:
(a) only present experimental data with no FEM modeling~\cite{Zhou2019-SaeIntJMaterManuf},~\cite{Gariya2022-MaterToday}, or
(b) only present FEM modeling with no experimental data~\cite{Shepherd2011-PNAS},~\cite{Sun2013-IROS},~\cite{Ellis2021-SoRo}, or
(c) only qualitatively compare the FEM and experimental data, and so the accuracy of their FEM models is not validated~\cite{Mosadegh2014-AdvFunctMater},~\cite{Martinez2012-AdvMater},~\cite{Zhang2020-JEngTechnolSci}.
In this work, for the first time, we present a comparison of FEM vs. experimental data for the sPN past 90$^{\circ}$; furthermore, we present this comparison over several cycles of bending.
We perform this comparison up to $\sim$170$^{\circ}$, and we demonstrate 96\% accuracy in the first trial (before fatigue accumulation).
Using this accurate model, we then analyze fatigue in a systematic manner.
For this, the paper leverages the use of a 3rd-order Yeoh model for Finite Element Analysis with coefficients determined by stress-strain experiments conducted up to 800\% strain~\cite{Kulkarni2015-Masters}.
We will then show that fatigue will result in a 20\% deviation from the ideal FEM behavior.
This observation highlights the importance and significance of considering fatigue in modeling soft robots and in the future for the control of such systems. 

The rest of this paper is organized as follows. 
In Section~\ref{sec.methods}, we present the end-to-end fabrication of the actuator, the FEM model, the pneumatic control of the actuator, and our automated data collection methods.
In Section~\ref{section.results}, we compare the simulated and experimental data, validating the 96\% accuracy of the FEM.
We then use this model to demonstrate and analyze the effect of fatigue in a targeted, precise manner.
In Section~\ref{section.conclusion}, we suggest how future researchers can use these results to conduct more informed analysis of soft actuator design and behavior.

\section{Methods}\label{sec.methods}


In this section, a brief overview of the fabrication of a soft actuator is given, followed by FEM modeling, pneumatic control, and data collection methods.
As discussed in the introduction, soft fluidic actuators made of elastomeric materials function by applying pneumatic or hydraulic pressure to the internal surface of a hyperelastic body.
The key is that the body is constrained on one or more surfaces by a flexible but inextensible strain-limiting layer.
When pressure is applied, the actuator only expands in the areas that are extensible.
The strain-limiting surfaces will bend but not elongate.
In the case of a pneu-net, the strain-limiting layer spans the entire base.
As the air chambers above the base expand, the limiting layer stops the actuator from elongating, which in turn causes bending.
Fig.~\ref{fig.sleeveFlextion} shows a fabricated sPN embedded into a wrist sleeve.
This is an early conceptual prototype to show how such an actuator could be used for wrist rehabilitation (flexion and extension).

\begin{figure}[htb] 
	\begin{center}
	\subfigure{
            \includegraphics*[width=1.0\columnwidth]
            {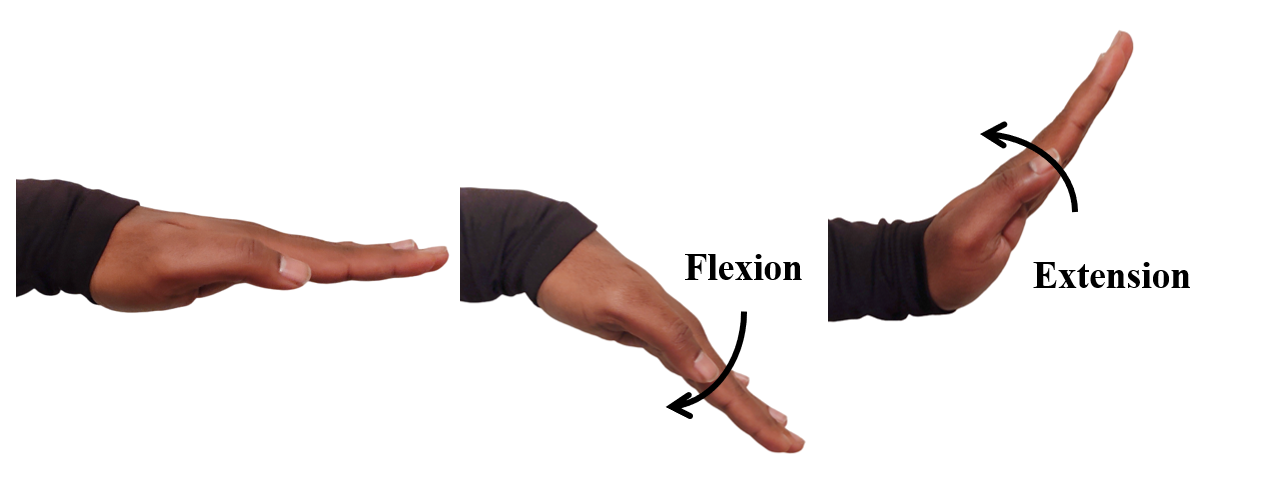}\label{subfig.wristFlexionExtension}
        }
	\subfigure{
            \includegraphics*[width=0.6\columnwidth]
            {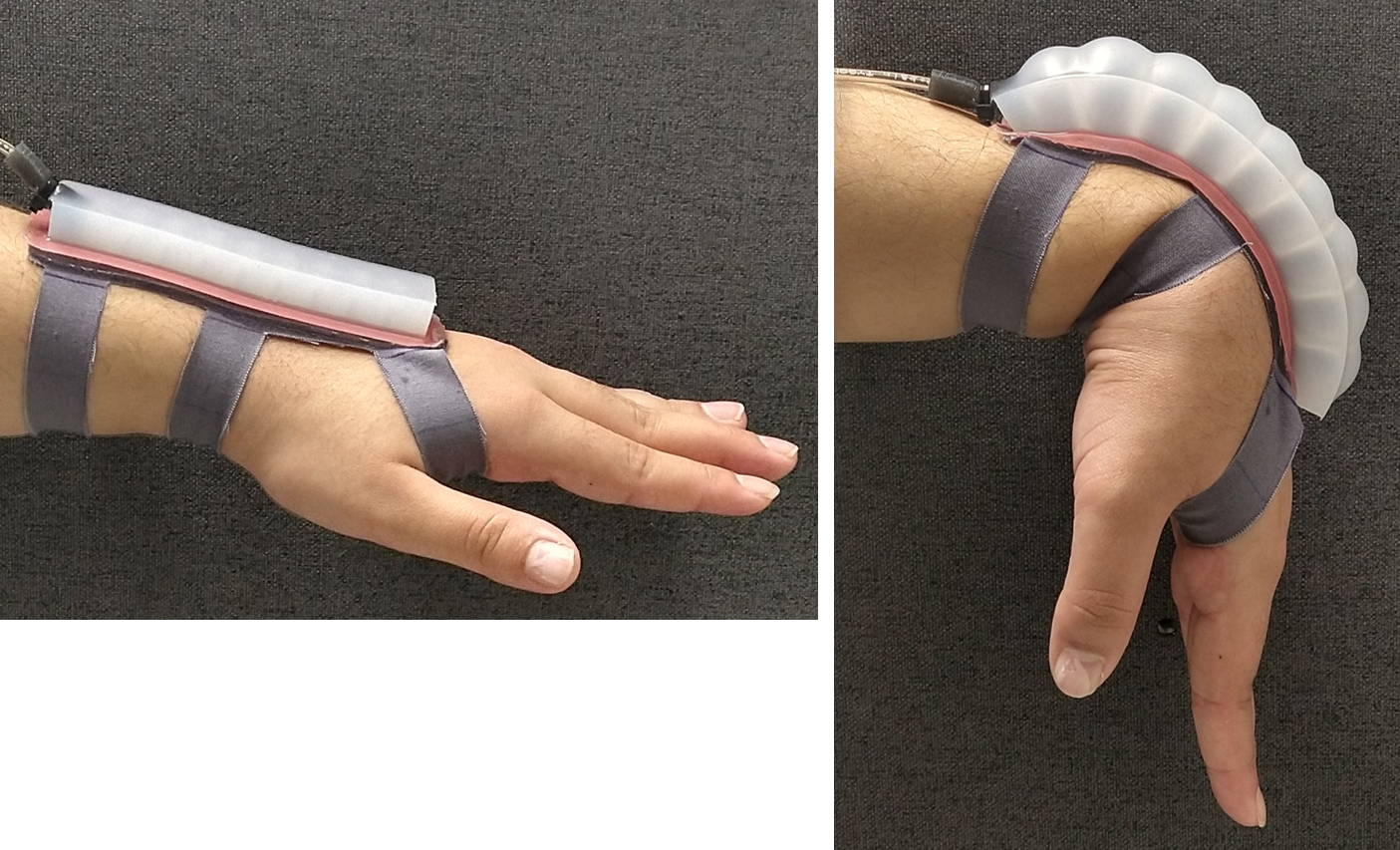}\label{subfig.sleeve}
        }        
	\caption{Early prototype of an sPN being used for wrist rehabilitation. \emph{Left}: Wrist motions: flexion and extension. emph{Right}: An sPN design embedded into an exo-sleeve for assisting wrist flexion or resisting wrist extension.}
	\label{fig.sleeveFlextion}
	\end{center}
\end{figure}


\subsection{Manufacturing}\label{subsection.manufacturing}

\begin{figure}[htb] 
	\begin{center}
	\subfigure[]{
            \includegraphics*[width=0.6\columnwidth]
            {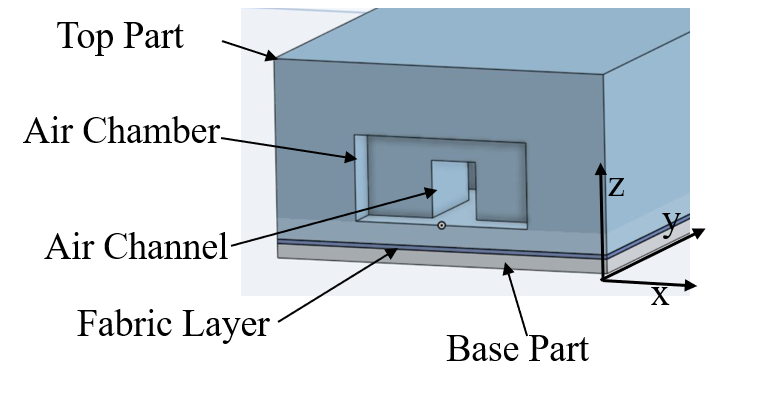}\label{subfig.section-x-z}
        }
	\subfigure[]{
            \includegraphics*[width=0.3\columnwidth]
            {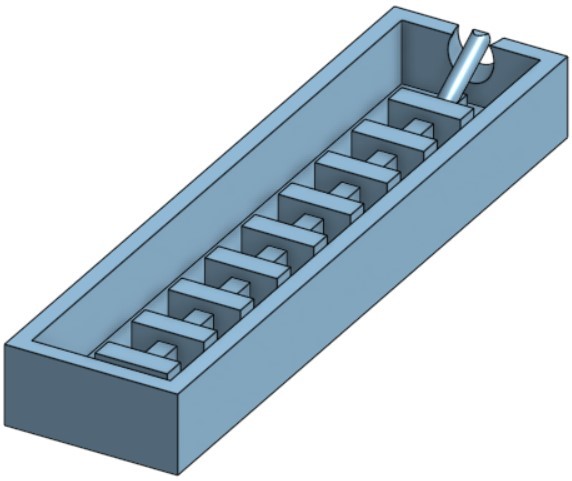}\label{subfig.mold}
        }        
	\caption{\subref{subfig.section-x-z} Transverse z-x plane cross-section of the CAD model for the actuator. The functional components of the actuator are displayed: the top part (or main body), the base, the limiting fabric layer, the air chambers, and the air channel. \subref{subfig.mold}: CAD model of the mold for the sPN top part (main body).}
	\label{fig.sleeveFlextion}
	\end{center}
\end{figure}

Fig.~\ref{subfig.section-x-z} shows a computer-aided design (CAD) model of a transverse cross-section of the sPN.
We define the y-axis as parallel to the main air channel, running along the length of the actuator.
The x-axis runs from side to side, and the z-axis runs from the bottom (the base layer) to the top.
This cross-section displays the components that allow the pneu-net to function.
The limiting fabric layer is embedded into the silicone base.
The top part, which is the main body, is sealed to the base.
The main body consists of internal air chambers, which are the nodes of the ``pneumatic network''.
Air is injected through the inlet into the channel, which then allows the chambers to expand against the limiting layer.
Note that the use of fabric for the limiting layer allows easy integration into clothing such as an exo-sleeve.

For manufacturing, silicone Ecoflex-50 (Smooth-On, Inc.) is used for the hyperelastic body, and a piece of inextensible cotton quilting fabric is used as the limiting layer.
Ecoflex-50 has a tensile strength of 315 psi (2.17 N/mm2) and can elongate up to 980\%, making it well-suited for this actuator, capable of handling large pressure inputs and capable of undergoing high deformations.
Fig.~\ref{subfig.mold} shows the CAD model for the mold for the top part, 3D printed on an Ultimaker 3.

\begin{figure}[htb]
    \begin{center}	
    \includegraphics*[width=1\columnwidth]{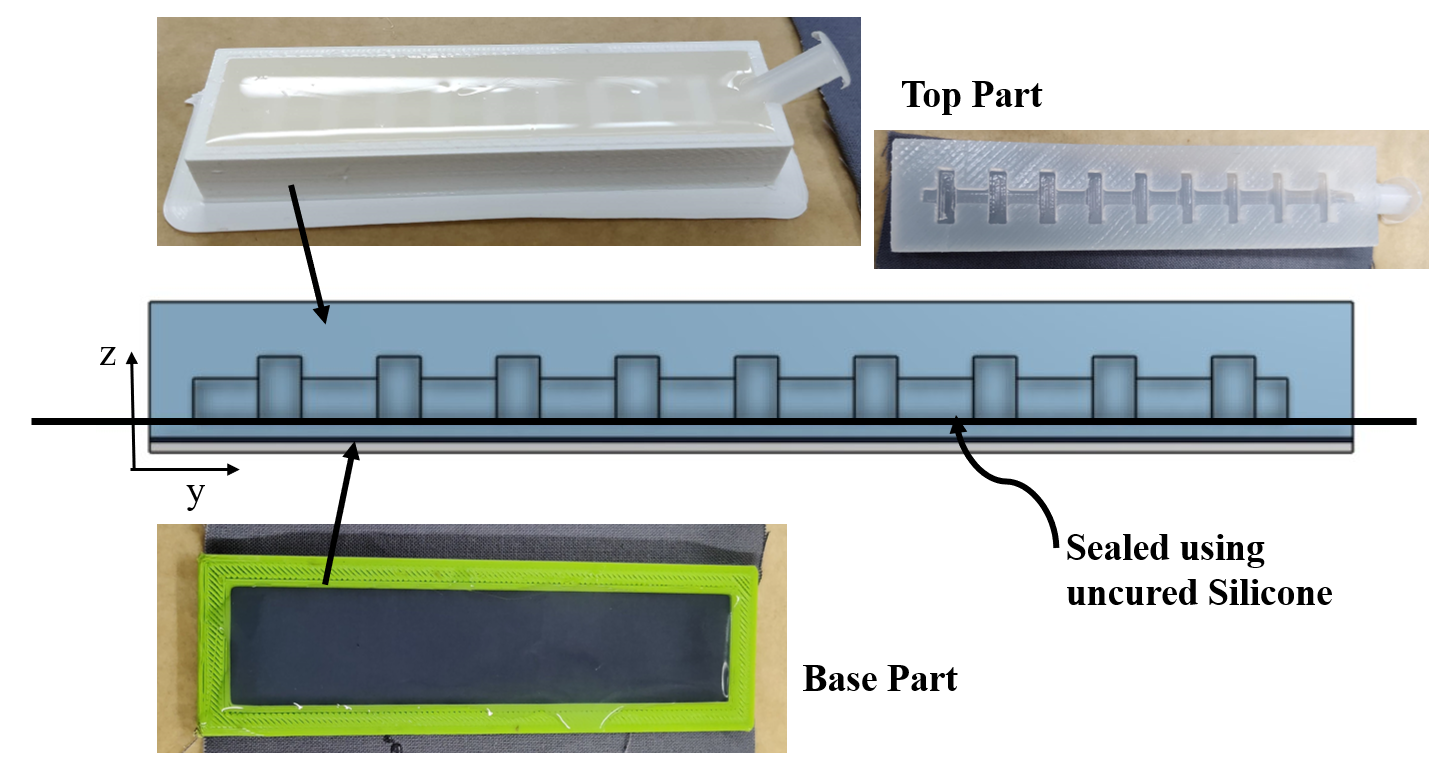}
    \caption[]{Manufacturing of an sPN. \emph{Top left}. The silicone is cast into the top mold, which is left to cure.\emph{Top right}. The top part (main body) after demolding. \emph{Middle}. y-z plane cross-section of the CAD model for the full actuator, showing a longitudinal side view. The thick black line is the sealing layer used to adhere the top to the base. \emph{Bottom} The silicone is cast into the base mold. The dark gray fabric is sandwiched between two thin layers of uncured silicone and then left to cure. The green frame is snapped on to help with a controlled fabrication.}
    \label{fig.manufacturing}
    \end{center}
\end{figure}

The Ecoflex-50 is mixed and degassed.
For the top mold, the mixture is poured and left to cure, as shown on the top left in Fig.~\ref{fig.manufacturing}.
The base mold is shown at the bottom of Fig.~\ref{fig.manufacturing}.
For the base mold, the mixture is first poured.
The fabric layer is then placed on the uncured silicone.
Finally, a very thin layer of silicone is poured on top of the fabric.

\begin{figure}[htb]
    \begin{center}	
    \includegraphics*[width=1\columnwidth]{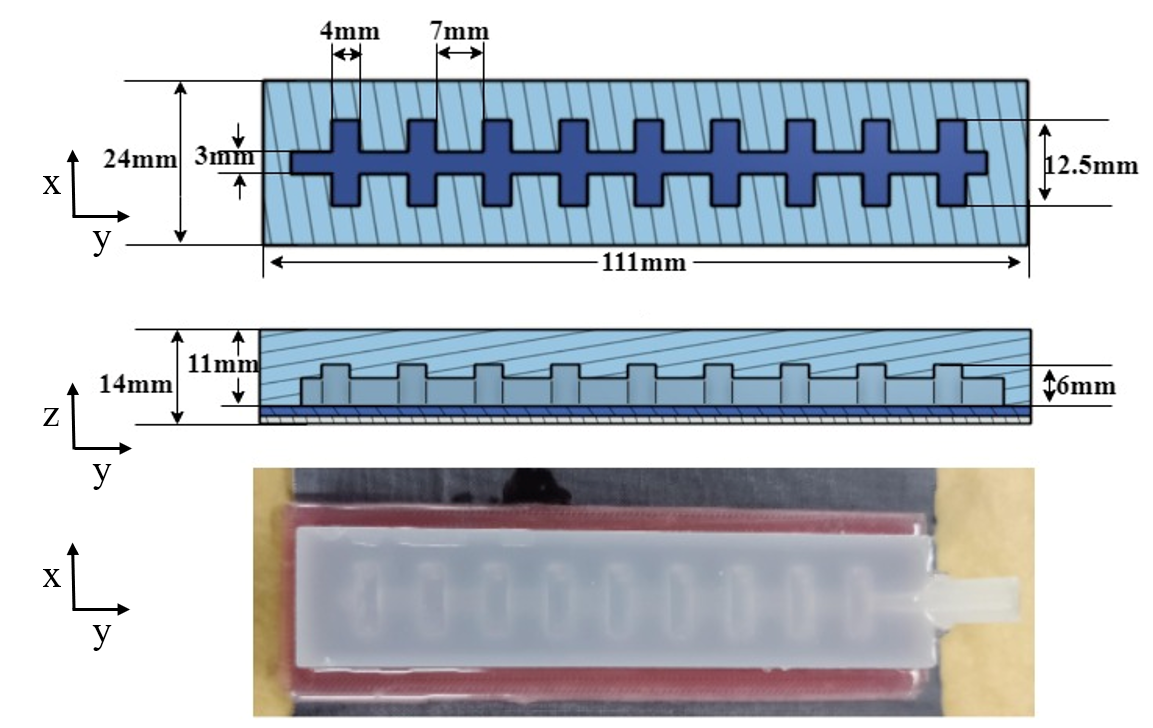}
    \caption[]{(Top) x-y plane cross-section of the CAD model, displaying a top view. (Middle) y-z plane cross-section of the CAD model, showing a longitudinal view. (Bottom) x-y plane top view of the manufactured actuator. The reddish-pink color is the pigmented sealing layer.}
    \label{fig.product}
    \end{center}
\end{figure}

After the top part is cured, it is demolded, as shown on the top right of Fig.~\ref{fig.manufacturing}.
A very thin ``sealing layer'' of uncured silicone is spread over the base part, and the top part is placed on it, as shown in the CAD diagram in the middle of Fig.~\ref{fig.manufacturing}.
The bottom of Fig.~\ref{fig.product} shows the final actuator fabricated by this process.
The top and middle of Fig.~\ref{fig.product} show the top and side CADs of the actuator, respectively, along with the dimensions chosen for all geometrical components.

\subsection{Finite Element Modeling (FEM)}\label{subsection.fem}

An FEM of the actuator design is constructed in order to compare the simulated and experimental static response of the actuator.
We define the static response as the bending angle that the actuator can achieve when a given pressure is applied at a steady state.
At the maximum bending angle, stress points were identified for fatigue analysis.
FEM is conducted in Abaqus FEA (Dassault Syst\`{e}mes) on CAD models generated with Onshape.
Each model is composed of 3 parts: the extensible top layer, the extensible sealing layer, and the inextensible limiting layer.
The extensible layers are characterized as isotropic hyperelastic solids, and the inextensible layer is characterized as linearly elastic.

In order to characterize hyperelastic silicone, polynomial curves are fit to the stress-strain data from material testing.
In general, four curve equations can be used: Neo-Hookean, Mooney-Rivlin, Ogden, and Yeoh.
The Ogden and Yeoh are higher-order models, which are recommended for strains above 400\%~\cite{marckmann2006comparison},~\cite{bhashyam2002ansys} and thus are the most appropriate for more compliant silicone rubbers such as Ecoflex-50 with no fiber reinforcement~\cite{Xavier2021-AdvIntellSyst}.
Stress-strain data should ideally be generated along multiple axes of tension; however, most studies just perform uniaxial testing due to testing equipment limitations~\cite{Xavier2021-AdvIntellSyst}.
Note that if only uniaxial testing is conducted, the Yeoh model will still perform well but the Ogden model will not~\cite{Xavier2021-AdvIntellSyst},~\cite{shahzad2015mechanical}.
A recent survey of FEM for soft actuators provides a comprehensive list of model coefficients for common silicones~\cite{Xavier2021-AdvIntellSyst}.
Of the experiments conducted for Ecoflex-50, they all perform uniaxial tests, and there is only one set of coefficients that are fit for a higher-order Yeoh model~\cite{Kulkarni2015-Masters}.
Furthermore, this is the only set that fits stress-strain data past 400\%; therefore, this is the model that we use.
This is a 3rd-order Yeoh model fit from experimental data up to 800\%.
We list these model coefficients in Table~\ref{table.kulkarniCoefficients}.
This material model can lead to error if used outside of the deformation range on which it was fit~\cite{marckmann2006comparison}, and in our experience, can lead to complete model failure as well.
The choice of this model for the hyperelastic material is the main reason for the success and accuracy of our FEM model for the sPN, which undergoes high deformation at high bending angles.
The rest of the FEM construction, as described below, is relatively standard.
 

\begin{table}[h]
	\renewcommand{\arraystretch}{1.3} 
	\caption[]{3rd-Order Yeoh model coefficients for Ecoflex-50~\cite{Kulkarni2015-Masters}}
	\centering
	\begin{tabular}{ l || l}
		\hline
		\textbf{Coefficient} & \textbf{Value} \\
		\hline
		\hline
		C$_{10}$ & 1.9 x 10$^2$  \\
		\hline
		C$_{20}$ & 9 x 10$^{-4}$ \\
		\hline
            \hline
		C$_{30}$ & -4.75 x 10$^{-6}$ \\
		\hline
	\end{tabular}
	\label{table.kulkarniCoefficients} 
\end{table}

The properties of the inextensible layer are set to be Young's Modulus = 6.5 GPa and Poisson’s ratio = 0.2.
These are the coefficients used in~\cite{Polygerinos2013-IROS} for modeling flexible and inextensible material.

In the Abaqus software, all components are merged, and then the model is meshed with a global seed size of 2.5mm and no local seeding.
When a merged part contains rigid and deformable components, local seeding can be employed to use a coarser mesh for the rigid components and a finer mesh for the deformable components, to optimize the tradeoff between resolution and computational efficiency.
The mesh is generated using tetrahedral elements for greater accuracy; free meshing is employed so that a finer mesh is generated in the boundary regions.
The geometric order for node placement is quadratic, and the mesh formulation is hybrid.
The fixed end-face of the actuator is defined with an encastre boundary condition to prevent translational and rotational displacements and velocities along all axes.
During the load step, pressure is applied uniformly to all internal surfaces of the air chambers and air channel.
The pressure is increased from 0 to 50 kPa using the default incrementation setting in Abaqus, with the procedure type set to ``static'' and ``general''.

\begin{figure}[htb]
    \begin{center}	
    \includegraphics*[width=1\columnwidth]{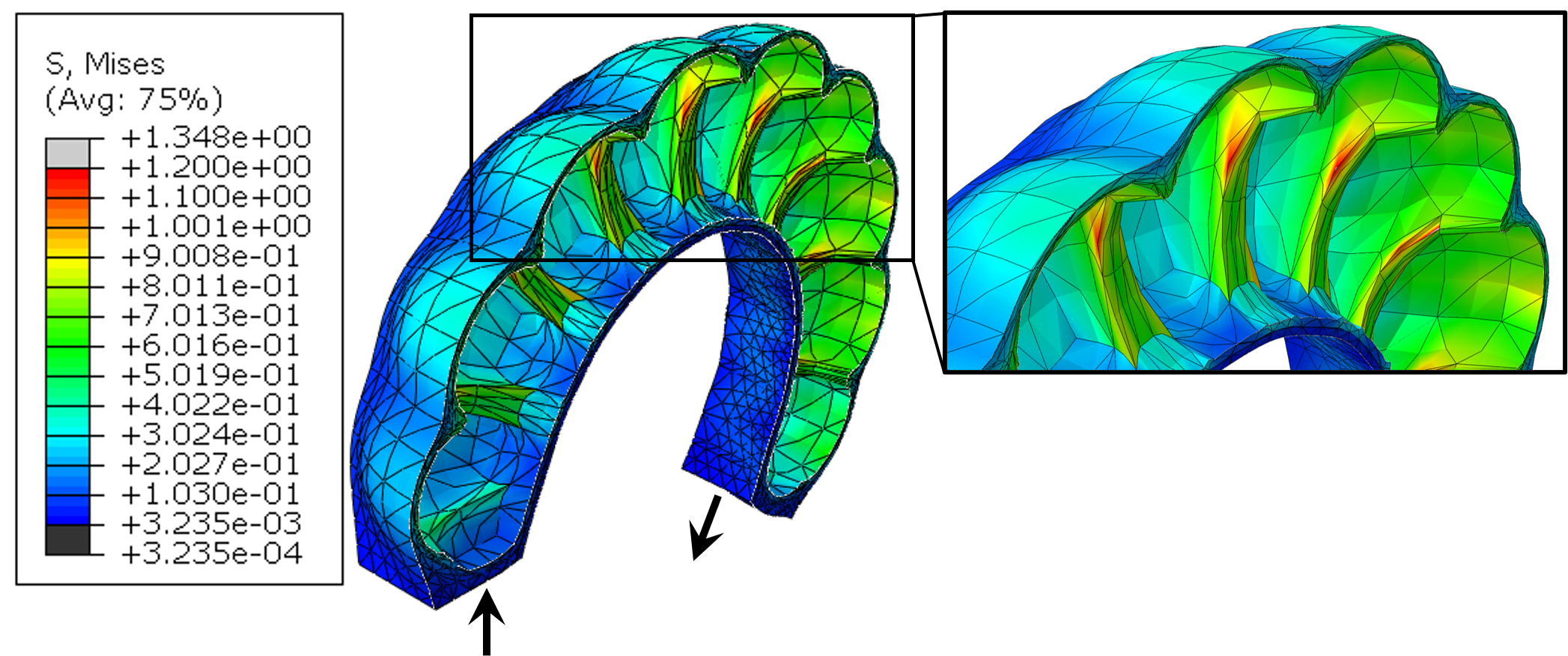}
    \caption[]{Equivalent stress (von Mises stress, unit: MPa) distribution at 50 kPa. \emph{Left:} A y-z plane cross-section of the actuator, showing a longitudinal side view. The bending angle is calculated between vectors shown with black arrows. \emph{Right:} Closeup of interior walls which experience the highest stress (in red).}
    \label{fig.fem}
    \end{center}
\end{figure}

Fig.~\ref{fig.fem} shows a y-z plane cross-section of the actuator's FEM pressurized to full 50 kPa, with the equivalent stress in MPa shown with the color bar.
The highest points of stress are shown in red, which occur on the interior walls, as would be expected for an sPN.
These high-stress regions are the most susceptible to fatigue, which can result in breaking and failure of the actuator under repeated bending.
Note that the compressive stresses on the inextensible layer will have larger magnitudes than the stresses on the extensible layers, so we remove the inextensible layer from the color-scale mapping in order to focus on the stress values in the deformable regions.
The bending angle is calculated between a vector normal to the fixed end-face and a vector normal to the moving end-face.
The vectors for the angle at 50 kPa are shown as black arrows in Fig.~\ref{fig.fem}.

\subsection{Data Collection and Control}\label{subsection.data}
In our experiments, the pressure required for actuation is provided by a pneumatic control station, which runs in MATLAB Simulink.
Realtime pressure was measured at the air inlet to the actuator, which was used as feedback for pressure control.
The pressure readings were sampled at 400 Hz.
	
\begin{figure}[htb]
    \begin{center}	
    \includegraphics*[width=.5\columnwidth]{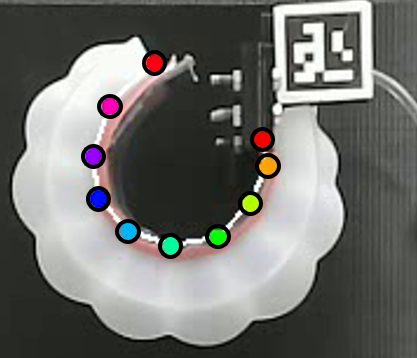}
    \caption[]{An sPN affixed to the test rig, pneumatically actuated with the control station. Markerless points track the limiting layer, which in turn is used to estimate the bending angle. The angle estimate at this image frame is 218$^{\circ}$.}
    \label{fig.dlc}
    \end{center}
\end{figure}

In order to track the behavior of the actuator, we utilized the DeepLabCut~\cite{Mathis2018-NatNeurosci} library, which is designed for tracking animal body parts.
The actuator was attached vertically to a test rig, and a stream of video images was logged at 30 fps, synchronized with the pressure readings.
Fig.~\ref{fig.dlc} shows a snapshot of the actuator attached to the test rig, actuated by the pneumatic control station, with tracked points from the DeepLabCut predictions.
The bending is considered to be the angle between the fixed end and the moving end.
(Multiple points were tracked to increase angle-measurement reliability.)

\section{Experiments and Results}\label{section.results}

\begin{figure*}[htb]
    \begin{center}	
    \includegraphics*[width=2\columnwidth]{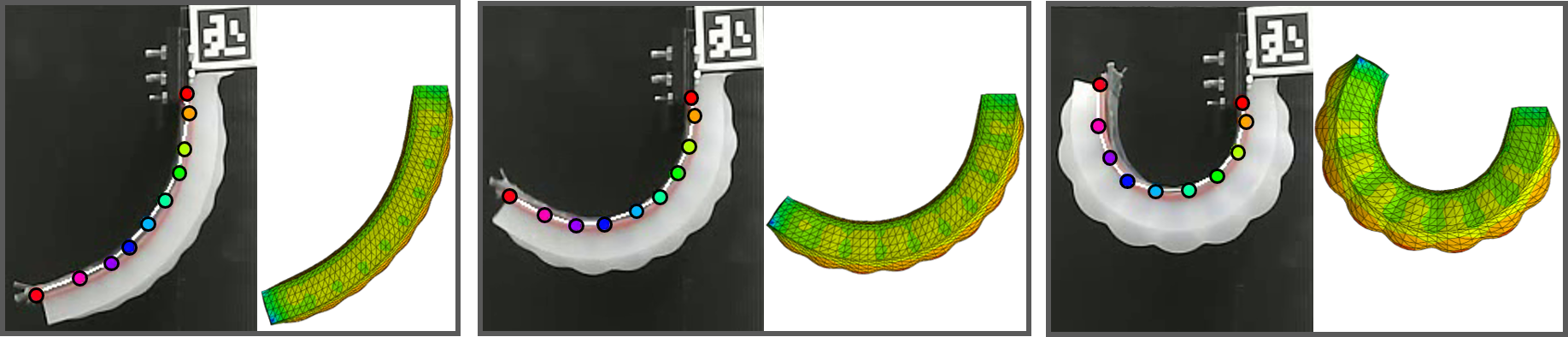}
    \caption[]{Side-by-side comparisons of the actuator and the FEM model. \emph{Left}, \emph{middle}, and \emph{right} views show angles of 73$^{\circ}$, 125$^{\circ}$, and 212$^{\circ}$, respectively. In each view, the right side shows the FEM model, and the left side shows the experimental behavior at the same angle.}
    \label{fig.sideBySides}
    \end{center}
\end{figure*}

In this section, we conduct experiments (static analysis) on bending angle response to pressure,
    compared to the behavior of the FEM, in order to:
    (a) validate the accuracy of our FEM model, and
    (b) analyze fatigue in the experimental data over repeated bending.
Fig.~\ref{fig.sideBySides} qualitatively depicts the ability of the simulation to correctly model the actuator's shape.
The left, middle, and right views show side-by-side comparisons of the FEM model and the pressurized actuator at angles of 73$^{\circ}$, 125$^{\circ}$, and 212$^{\circ}$, respectively.

\begin{figure}[htb] 
	\begin{center}
	\subfigure[Trial 1, Time Series, Experimental]{
            \includegraphics*[width=1.0\columnwidth]
            {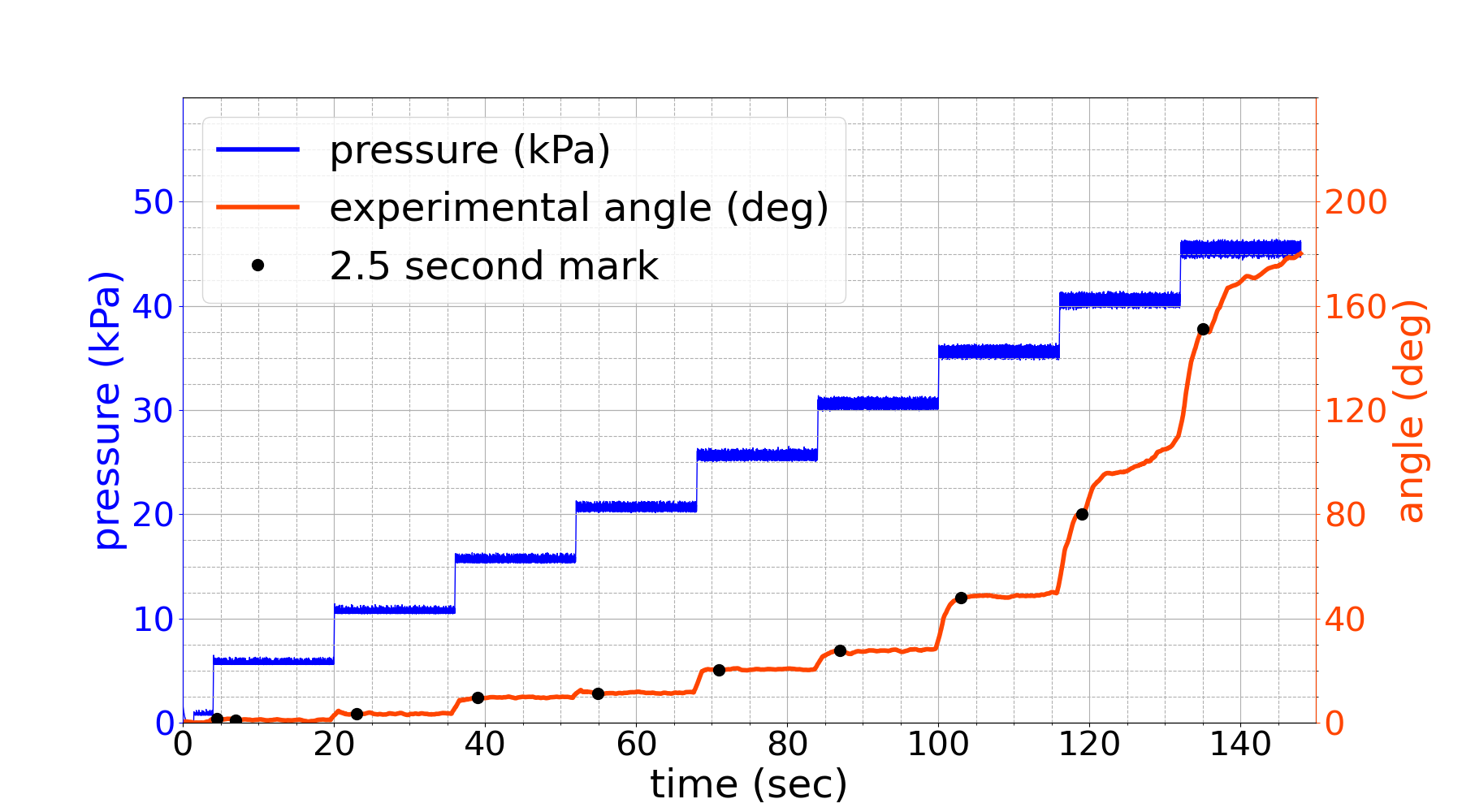}\label{subfig.anglePressureVsTime01}
        }
	\subfigure[Trial 1, Angle vs Time, FEM and Experimental]{
            \includegraphics*[width=1.0\columnwidth]
            {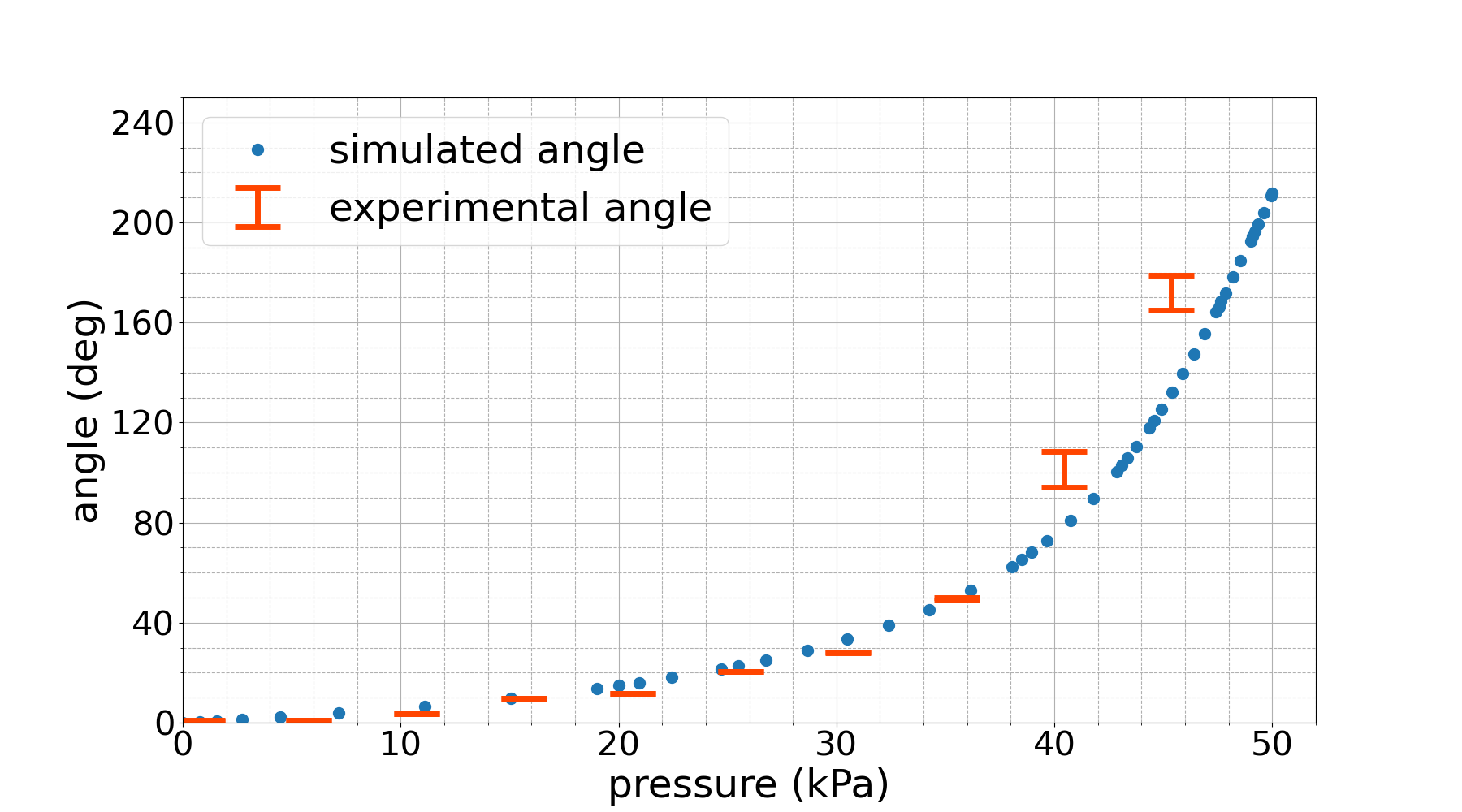}\label{subfig.angleVsPressure01}
        }
	\caption{
            Experimental data from the first trial.  
            \subref{subfig.anglePressureVsTime01} The blue data points are the pressure readings of the staircase function. The orange data points are the angle measurements.
            \subref{subfig.angleVsPressure01} Angle vs. pressure plot of experimental and FEM data. The blue data points are from the FEM model. The orange distribution bars are the experimental data. Each bar is one of the staircase steps.
            }
	\label{fig.trial01}
	\end{center}
\end{figure}

In order to quantitatively analyze the performance of the system with that of the simulation, a series of fixed pressures were applied to the actuator, and the resulting error was examined at steady-state.
Each step of the staircase pressure function holds the pressure constant for 16 seconds.
Successive steps increment by 5 kPa, ranging from 0 to 45 kPa.
The purpose of the staircase function is to allow the actuator to reach a steady state at each pressure so that the measurements can be compared to the static FEM model.
Fig.~\ref{fig.trial01} shows data collected from an actuator on its first trial after fabrication.
Fig.~\ref{subfig.anglePressureVsTime01} shows a Y-Y plot of the time series data.
The blue data points are the readings from a pressure sensor located at the air inlet to the actuator.
The resulting bending angle measurements from the markerless computer vision system are shown in orange.
We observe that the actuator is able to fully respond to the pressure increment and reach a steady state in most cases within 2.5 seconds past the start of the pressure steps.
This 2.5-second mark is shown with black points overlaid on the orange curve.
We repeated this staircase experiment for ten trials (each with ten steps from 0 to 45 kPa for 16 seconds) to induce fatigue.

In Fig.~\ref{subfig.angleVsPressure01}, we compare our experimental data from trial one to the simulated data from our FEM model.
The angle measurements from each 5kPa step in the staircase are displayed here in orange as a distribution from 2.5 seconds and after.
The bottom of the orange distribution is the angle at the 2.5-second mark corresponding to the black point in Fig.~\ref{subfig.anglePressureVsTime01}.
The top of the orange distribution is the end of the corresponding pressure step (at 16 seconds when the angle is also the highest).
For lower pressures where the angle plateaus, the bottom and top angles are the same; thus, the distribution appears as a flat line.
The distribution becomes wide at the higher-pressure steps that do not plateau.
No matter the size of the distribution, the bottom (the 2.5-second mark) is very close to the FEM data for all pressure steps, demonstrating the ability of the simulation to accurately model the actuator.
 
\begin{figure}[htb] 
	\begin{center}
	\subfigure[Trial 2, Angle vs Time, FEM and Experimental]{\includegraphics*[width=1\columnwidth]
        {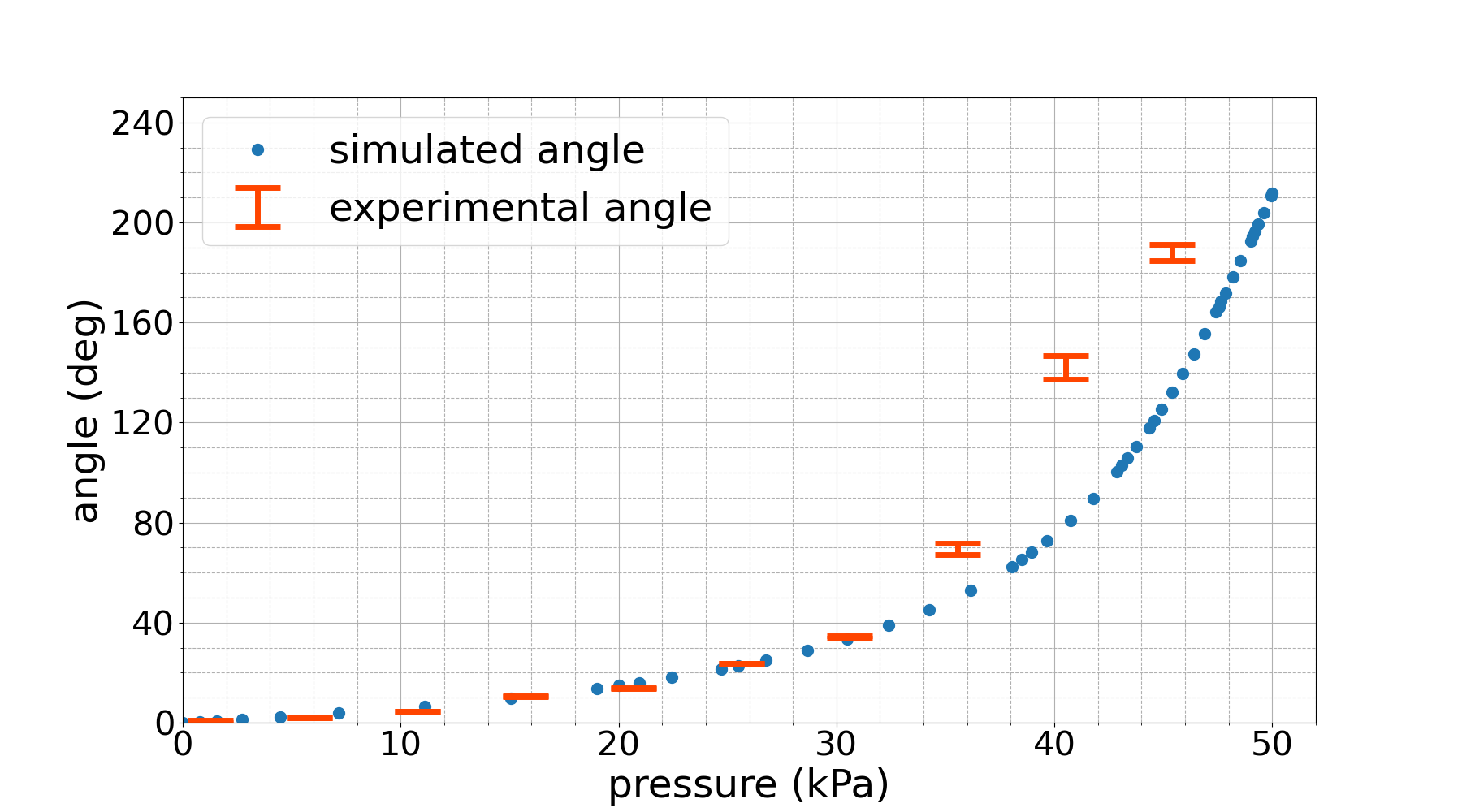}\label{subfig.angleVsPressure02}}
	\subfigure[Trial 10, Angle vs Time, FEM and Experimental]{\includegraphics*[width=1\columnwidth]
        {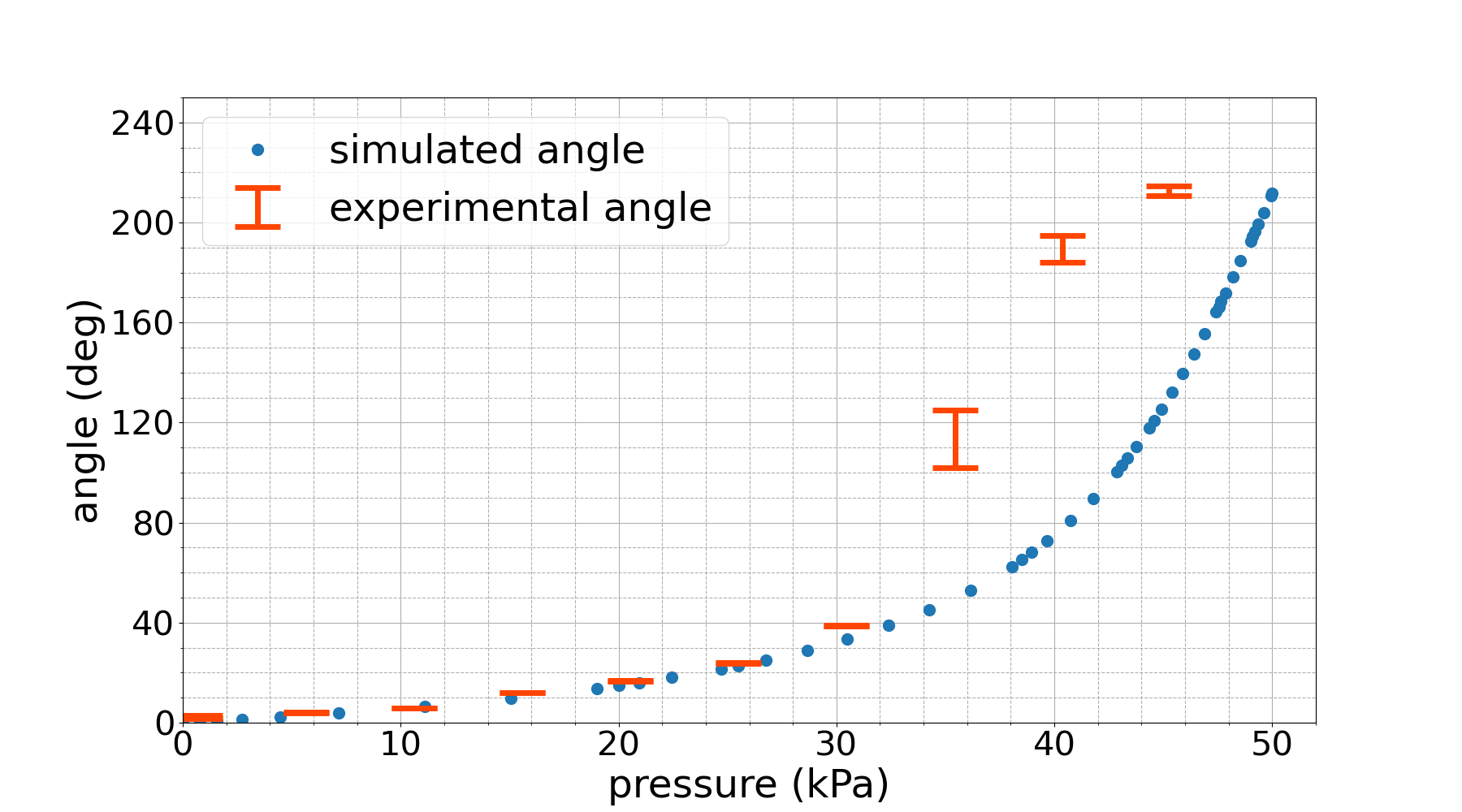}\label{subfig.angleVsPressure10}} 
	\caption{Subsequent trials after the first trial.  Angle vs. pressure plots of experimental and simulated data, similar in format to Fig.~\ref{subfig.angleVsPressure01}. Trial 2 shows the immediate effect after one use.  Trial 10 shows the effect after repeated rounds of bending.}
	\label{fig.trialMult}
    \vspace{-4.00mm} 
	\end{center}
\end{figure}

Fig.~\ref{fig.trialMult} shows subsequent data collection trials with the same staircase reference signal.
Fig.~\ref{subfig.angleVsPressure02} is the trial directly after the first one, and we see that for the two highest pressures, the bottom of the distribution is no longer close to the simulation, and we also see that the third highest pressure is starting to show non-plateau behavior.
This is caused by the micro-fractures in the body of the system due to fatigue, which changes the behavior of the system and causes the system to deviate from the ideal FEM model.
Fig.~\ref{subfig.angleVsPressure10} is after ten trials.
We see that the three highest pressures have risen above the simulation with non-plateau behavior, while the lower pressures continue to match the simulation and reach steady-state behavior.

\begin{figure}[htb]
    \begin{center}	
    \includegraphics*[width=1\columnwidth]{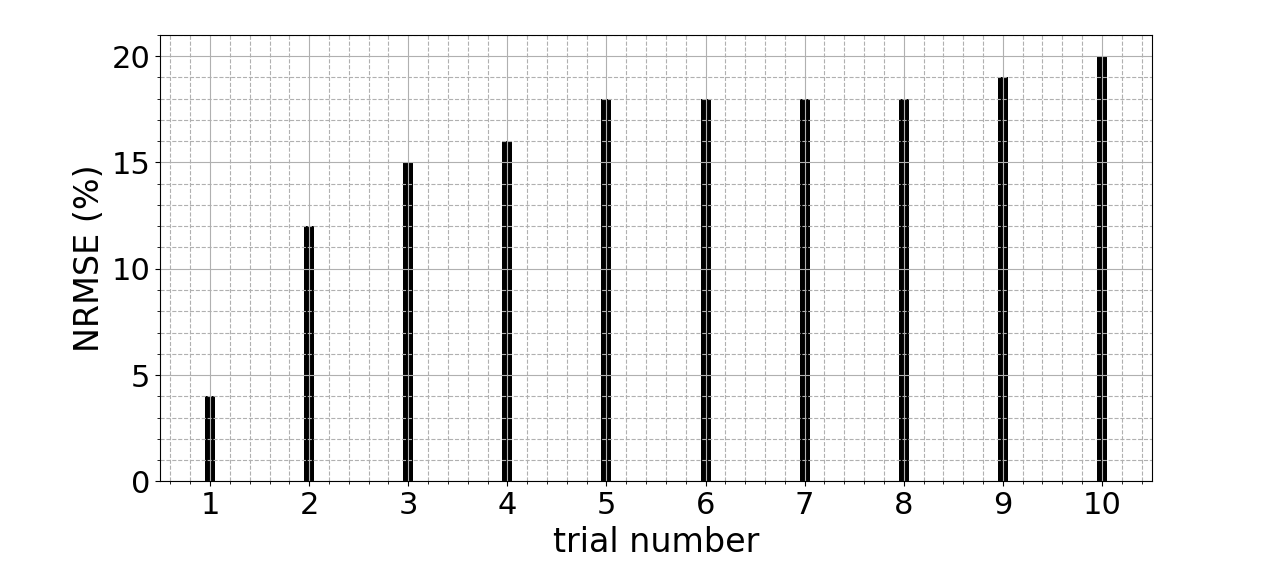}
    \caption[]{NRMSE of each trial. The error is calculated between the FEM and the experimental bending angles over all 5 kPa pressure steps.}
    \label{fig.nrmse}
    \vspace{-4.00mm} 
    \end{center}
\end{figure}

Fig.~\ref{fig.nrmse} shows the normalized RMSE (NRMSE), highlighting the deviation (the error) from the FEM model.
Each bar is for a corresponding trial.
Computing the NRMSE allows us to compare experimental and simulated data.
It can be seen that the error for trial one is 4\%, and for trial 10 it goes up to 20\% as a result of fatigue.
This highlights the major effect of fatigue on the behavior of the system.


\begin{figure}[htb]
    \begin{center}	
    \includegraphics*[width=1\columnwidth]{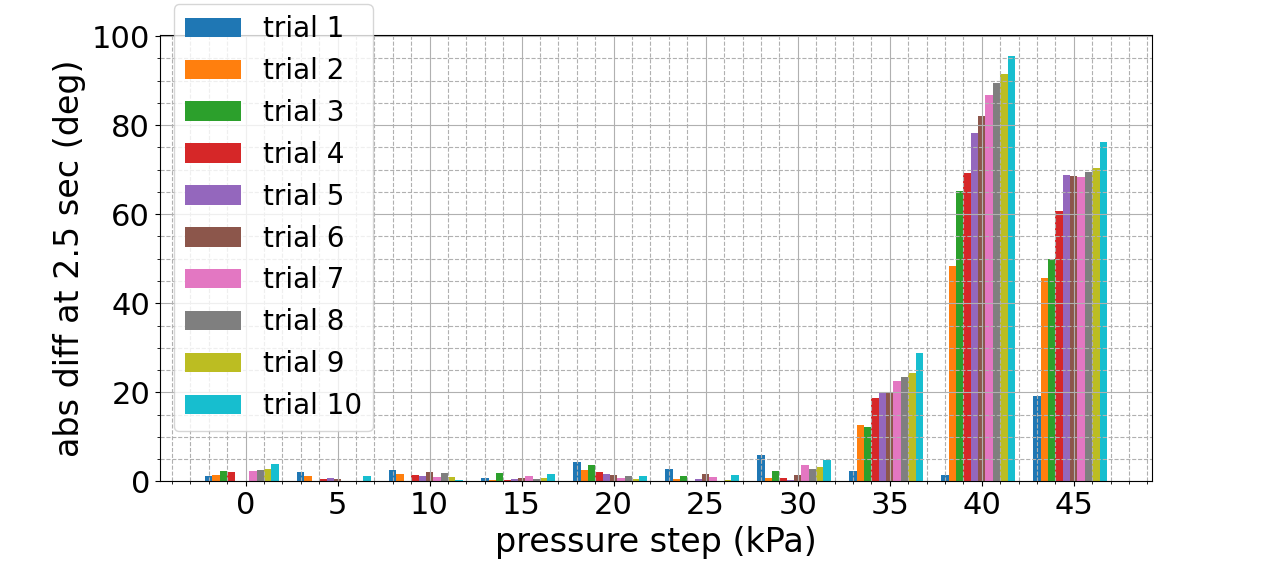}
    \caption[]{Bar plot of the error of each trial at each 5 kPa step. The main groups on the x-axis are the pressure steps. Each color within each group is for a particular trial.}
    \label{fig.absErr}
    \vspace{-4.00mm} 
    \end{center}
\end{figure}

Fig.~\ref{fig.absErr} characterizes the error for each pressure step.
Each bar in the bar plot is for one 5 kPa step and one trial.
The height of the bar is the error, which is the absolute value of the difference between the angle predicted by the simulation at that pressure and the actual angle at that pressure after it has been at that pressure step for 2.5 seconds.
Each pressure step is shown as a main group of bars on the x-axis, and each trial is a different color within each main group.

\begin{figure}[htb]
    \begin{center}	
    \includegraphics*[width=1\columnwidth]{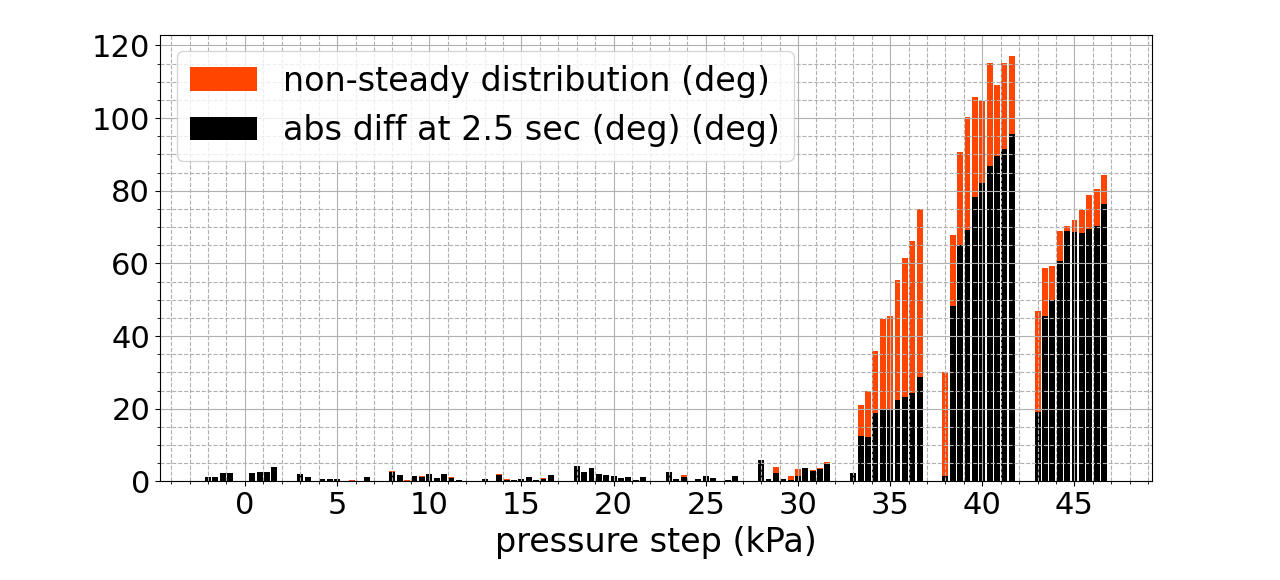}
    \caption[]{Bar plot of the error of each trial at each 5 kPa step. The main groups on the x-axis are the pressure steps. Each bar within each group is for a particular trial. The colored bars from Fig.~\ref{fig.absErr} are now shown in black. The height of the black portion of the bar is the error between the FEM and the angle at 2.5 seconds. The height of the orange portion is the angle at 16 seconds.}
    \label{fig.absDoubleErr}
    \vspace{-4.00mm}
    \end{center}
\end{figure}

Fig.~\ref{fig.absDoubleErr} shows the same organization of steps and trials
	as in Fig.~\ref{fig.absErr},
	but now includes the distribution of angle values after 2.5 seconds.
The colored bars from Fig.~\ref{fig.absErr} are now simplified to be all in black,
	and the distribution from 2.5 seconds to 16 seconds is plotted above each bar in orange.
The orange portion of the bar corresponds to the orange distributions in the previous Angle vs. Pressure graphs.

In Fig.~\ref{fig.nrmse}, we observe that the first trial exhibits a very low NRMSE error of 4\% with respect to the simulated model, but then the error increases significantly.
Investigating further into subsequent trials, we observe in Fig.~\ref{fig.absErr} that for all but the three highest pressures, the error stays very low for all ten trials.
In Fig.~\ref{fig.absErr}, we observe that for the highest two pressures, the error significantly jumps up after the first trial.
Looking at this trend in Fig.~\ref{fig.absDoubleErr}, we notice that there is a clear correlation between the presence of error from the simulation and non-steady state behavior.
This non-steady state behavior is due to the forming of microtears in the silicone, resulting in fatigue.
This fatigue then changes the actuator's behavior, making it deviate from the simulated model.

\section{Conclusion}\label{section.conclusion}

Pneumatic soft actuators undergo high deformation, making them hard to model even without fatigue.
To the best of the authors' knowledge, this work is the first to show an FEM vs. experimental comparison for an sPN with angles exceeding 90$^{\circ}$.
We have shown that our FEM model can predict the bending angle of a newly-fabricated sPN up to 167$^{\circ}$; furthermore, this prediction has a very low error of 4\% before repeated use.
Capitalizing on the accuracy of the FEM model, this paper focuses on the effect of fatigue over repeated use of the actuator. We have shown that fatigue can cause significant errors in the behavior of the actuator deviating from the ideal (FEM-based) prediction.  Using computer vision, the bending angle was captured with fine temporal resolution, demystifying the dynamic step response, which allowed the observation of a correlated pattern between deviation from the FEM model and the non-steady-state behavior.

This paper highlights the importance and significance of taking into account the fatigue behavior caused by micro-fractures, which can later result in macro-fractures and actuator failure. 
In the future, the outcomes from this study can be used to (a) select more durable materials with lower fatigue, (b) optimize geometrical designs to minimize fatigue, (c) guide the bending angle thresholds given fatigue provocation, and (d) oversee seamless soft-robot control through objective time-variable fatigue monitoring. 
\clearpage
                                  
\bibliographystyle{./IEEEtran}
\bibliography{./IEEEabrv,./references}

\end{document}